\pdfoutput=1

\documentclass[11pt]{article}

\usepackage{acl}

\usepackage{times,latexsym}
\usepackage{url}
\usepackage{amsfonts}

\usepackage{times}
\usepackage{setspace}
\usepackage{booktabs,siunitx}
\usepackage{multirow}
\usepackage{latexsym}
\usepackage{hyperref}
\usepackage{cleveref}
\usepackage{xcolor, colortbl}
\usepackage{xspace,mfirstuc,tabulary}
\usepackage{url}
\usepackage{comment}
\usepackage{makecell}
\usepackage{verbatim}
\usepackage{latexsym}
\usepackage{enumitem}
\usepackage{caption}
\usepackage{subcaption}
\usepackage{float}

\usepackage{graphicx}
\usepackage{tikz}
\usepackage[normalem]{ulem}
\usepackage[T1]{fontenc}
\usepackage{soul} 
\usepackage[utf8]{inputenc}

\usepackage{microtype}

\usepackage{inconsolata}

\title{The Bias Amplification Paradox in Text-to-Image Generation}

\author{Preethi Seshadri \\
UC Irvine \\
  \texttt{preethis@uci.edu}\\\And
  Sameer Singh \\
  UC Irvine \\
  \texttt{sameer@uci.edu}\\\And
  Yanai Elazar \\
  Allen Institute for AI \\ University of Washington \\
  \texttt{yanaiela@gmail.com}}

\begin{document}
\maketitle
\begin{abstract}
Bias amplification is a phenomenon in which models exacerbate biases or stereotypes present in the training data.
In this paper, we study bias amplification in the text-to-image domain using Stable Diffusion by comparing gender ratios in training vs. generated images. 
We find that the model appears to amplify gender-occupation biases found in the training data (LAION) considerably. 
However, we discover that amplification can be largely attributed to discrepancies between training captions and model prompts.
For example, an inherent difference is that captions from the training data often contain explicit gender information while our prompts do not, which leads to a distribution shift and consequently inflates bias measures.
Once we account for distributional differences between texts used for training and generation when evaluating amplification, we observe that amplification decreases drastically.
Our findings illustrate the challenges of comparing biases in models and their training data, and highlight confounding factors that impact analyses.\footnote{We release the code at: \url{https://github.com/preethiseshadri518/bias-amplification-paradox/}} 
\end{abstract}

\section{Introduction}

Breakthroughs in machine learning have been fueled in large part by training models on massive unlabeled datasets \citep{gao2020pile, 2020t5, LAION}.  
However, several studies have shown that these datasets exhibit biases and undesirable stereotypes \citep{birhane-et-al2021, dodge-documenting, garcia2023uncurated}, which in turn impact model behavior.
Given that models are trained to represent the data distribution, it is not surprising that models perpetuate biases found in the training data \citep[among others]{bias-in-bios, sap-etal-2019-risk, Adam_2022}. 

\begin{figure}[t]
    \centering
    \includegraphics[width=\linewidth]{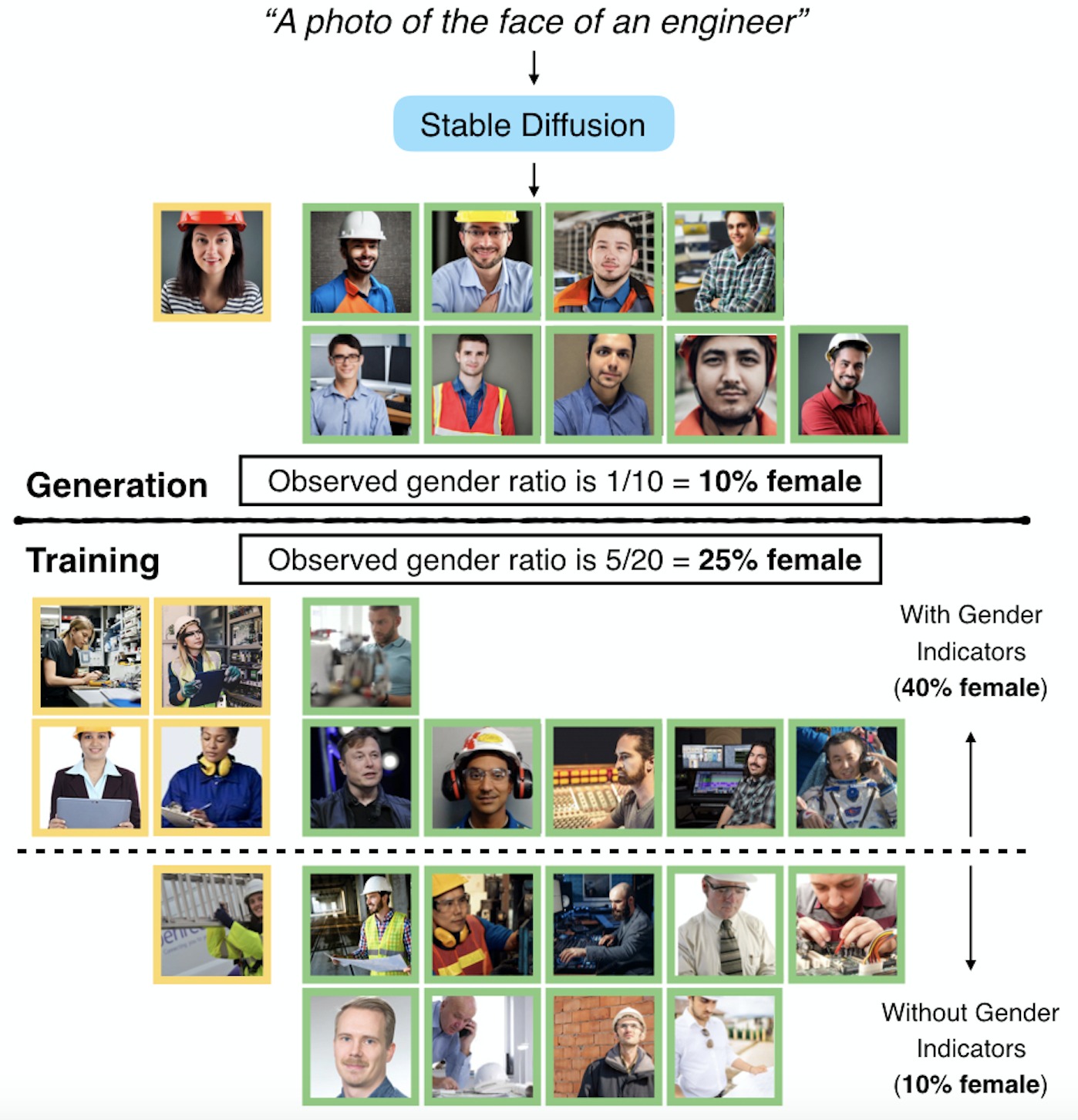}
    \caption{Comparing generated and training images for \textbf{engineer}, the model clearly seems to amplify bias by going from 25\% to 10\% female in training vs. generated images. However, when looking at the subset of training examples \textit{without gender indicators} in captions (similar to our prompts), the model does not amplify bias.}
    \vspace{-1.0em}
    \label{fig:intro}
\end{figure}

Imagine that a model generates images of engineers that are female 10\% of the time. 
When examining the training data, we may assume that models reflect associations in the data and expect to observe roughly 10\% female as well.\footnote{Note that even such bias preservation may be undesirable.} 
However, it would be problematic for a model to instead exacerbate existing imbalances by generating engineer images that are only 10\% female, while the training engineer images are 25\% female, as shown in Figure \ref{fig:intro}.
This phenomenon, known as \textit{bias amplification} \cite{zhao-etal-2017-men}, is concerning because it further reinforces stereotypes and widens disparities.
While previous works suggest that models amplify biases \citep{zhao-etal-2017-men, wang2019iccv, hall-et-al2022, hirota-et-al2022, friedrich2023FairDiffusion}, there remain unanswered questions about the paradoxical nature of bias amplification: \textit{Given that models learn to fit the training data, why do models intensify biases found in the data as opposed to strictly representing them?}

In this paper, we investigate how model biases compare with biases found in the training data. 
We focus on the text-to-image domain and analyze gender-occupation biases in Stable Diffusion, \citep{rombach-et-al2022} as well as its publicly available training dataset LAION \citep{LAION}, which consists of image-caption pairs in English (\S \ref{sec:setup}).
To select training examples, we identify captions that mention an occupation (e.g., engineer) and obtain corresponding images. 
We follow previous work \citep{bianchi-et-al2022, luccioni2023stable} and use prompts that contain a given occupation (e.g., ``A photo of the face of an engineer'') to generate images. 
For each occupation, we then classify binary gender to measure bias in corresponding training and generated images, and compare the respective quantities to determine whether the model amplifies biases\footnote{We define bias as a deviation from the 50\% balanced (binary) gender ratio. This definition differs from measuring performance gaps between groups (e.g., TPR difference), which is common in classification setups.} from its training data (\S \ref{sec:methodology}).

At first glance, it appears that the model amplifies bias considerably (12.57\%) using existing approaches (\S \ref{sec:keyword-querying}). 
However, we discover clear distributional differences when comparing how training captions and prompts are written, which impact amplification measurements. 
For example, an inherent distinction is that captions often contain explicit gender information while prompts used to study gender-occupation biases do not.\footnote{Since we study gender bias, prompts exclude explicit gender information to avoid skewing generations.}
As shown in Figure \ref{fig:intro}, the gender distribution for captions with gender indicators (40\% female) clearly differs from the distribution for captions without such indicators (10\% female) for the occupation engineer.

Instead of naively considering all training captions that contain a given occupation when analyzing bias amplification, we propose focusing on subsets of the training data that reduce distribution shifts between training and generation (\S \ref{sec:reducing-discrepancies}).
We introduce two approaches to account for distributional differences observed in qualitative evaluation: (1) Excluding captions with explicit gender information and (2) Using nearest neighbors (\textsc{NN}) on text embeddings to select training captions that closely resemble prompts. 
Both approaches restrict the search space of training texts to more closely match prompts, which results in considerably lower amplification measures. 
We then eliminate differences between training captions and prompts by utilizing the captions themselves to generate images (\S \ref{sec:captions-as-prompts}), and show that amplification is minimal. 
By modifying either the captions or prompts used to evaluate amplification, we provide insights into how the subsets of data used to measure bias at training and generation impact amplification. 

To summarize, we study gender-occupation bias amplification in Stable Diffusion and highlight notable discrepancies between texts used for training and generation.
We demonstrate that naively quantifying bias provides an incomplete and misleading depiction of model behavior. 
Our work emphasizes that comparisons of dataset and model biases should factor in distributional differences and evaluate comparable distributions.
We hope that our work encourages future studies that analyze model behavior through the lens of the data.

\section{Experimental Setup}\label{sec:setup}
\subsection{Dataset and Models}
To study bias amplification, we use Stable Diffusion \citep{rombach-et-al2022}, a text-to-image model that generates images based on a textual description (prompt). 
Stable Diffusion is trained on pairs of captions and images taken from LAION-5B \citep{LAION}, a public dataset created by scraping images and their captions from the web. 
We focus on two versions, Stable Diffusion 1.4 and 1.5, which are both trained on text-image pairs from the 2.3 billion English portion of LAION-5B.\footnote{Stable Diffusion 1.5 is finetuned for a longer duration on LAION-Aesthetics (a subset of higher quality images).}


\subsection{Gender Classification}
We analyze bias in images with respect to perceived gender.\footnote{Classifying binary gender based on appearance has limitations and perpetuates stereotypes. While our analysis excludes non-binary individuals, inferring non-binary gender from appearance alone risks misrepresenting a marginalized group.} 
To classify binary gender at scale, we utilize an automated approach.
Therefore, it is important to verify that images include faces, and that perceived gender is discernible from these images. 
We first check whether an image contains a single face using a face detector.\footnote{\url{https://developers.google.com/mediapipe/solutions/vision/face_detector/python}} 
Then, we use CLIP \citep{Radford2021LearningTV}, a multimodal model with zero-shot image classification capabilities, to predict gender (note that Stable Diffusion also uses CLIP's text encoder to encode prompts).
To exclude cases where gender is difficult to infer (e.g., faces might be blurred or obscured), we only consider images for which the predicted gender probability is greater than or equal to 0.9. 
We apply these filtering steps to training and generated images.

\subsection{Occupations}
Similar to previous works, we analyze gender-occupation biases for occupations that exhibit varying levels of bias \citep{rudinger-etal-2018-gender, zhao-etal-2018-gender, bias-in-bios}. 
These include occupations that skew male (e.g., CEO, engineer), fairly balanced (e.g., attorney, journalist), and female (e.g., dietitian, receptionist) based on the training data. 
In total, we consider 62 job occupations, which can be found in Table \ref{tab:occs} in the Appendix.

\section{Methodology}\label{sec:methodology}

\begin{table}[t]
 \centering
 \small
  \begin{tabular}{ll}
    \toprule        
    \bf \# & \bf Prompt \\
    \midrule
    1 & A photo of the face of a/an [OCCUPATION] \\
    2 & A portrait photo of a/an [OCCUPATION] \\
    3 & A photo of a/an [OCCUPATION] smiling \\
    4 & A photo of a/an [OCCUPATION] at work \\
    \bottomrule
  \end{tabular}
   \caption{The four prompts we use to generate images. ``[OCCUPATION]'' is a placeholder we replace with one of the 62 occupations we use.}
   \vspace{-1.0em}
  \label{tab:prompts}
\end{table}

\subsection{Measuring Model Bias}
\label{sec:mod_bias}

To measure biases exhibited by the model, we generate images using four prompts, shown in Table \ref{tab:prompts}. 
These prompts deliberately do not contain gender information since we want to capture biases learned by the model.
Both prompts \#1 and \#2 also direct the model to generate faces by including ``face''/``portrait''. 
We generate 500 images per occupation and prompt using various random seeds (which is used to generate random noise).
We define $G_{P_o}$ as the percentage of females in generated images for a prompt $P$ describing an occupation $o$.

\subsection{Measuring Data Bias}
\label{sec:tr_bias}
Given that the training data consists of image-caption pairs, we use captions to obtain relevant training examples. 
In doing so, we assume that the training captions relating to a given occupation mention the occupation.
We use the search capabilities of WIMBD \citep{elazar2023whats}, a tool that enables exploration of large text corpora, to query LAION. 
We define $T_{S_o}$ as the percentage of females in images for a training subset $S$ corresponding to occupation $o$ (we provide more details on example selection in Section \ref{sec:keyword-querying}).

\subsection{Evaluating Bias Amplification}
We compute bias amplification by comparing the percentage female in generated ($G_{P_o}$) vs. training ($T_{S_o}$) images for a specific occupation $o$ using the approach outlined in \citet{zhao-etal-2017-men}:
$$ A_{P_o,S_o} = |G_{P_o}-50| - |T_{S_o}-50|$$

This formulation takes into account that amplification for a given occupation is specific to the prompt $P_o$ used to generate images, as well as the chosen subset of training examples $S_o$. 
For a set of occupations $O$, the expected amplification is: 

$$\mathop{\mathbb{E}}_{o \in O} [A_{P_o,S_o}] = \frac{1}{|O|}\sum_{o \in O}A_{P_o,S_o}$$

$A_{P_o, S_o}$ is calculated for each occupation and aggregated across occupations ($O$) to obtain $\mathbb{E}[A_{P_o,S_o}]$ for each prompt.
We then average $\mathbb{E}[A_{P_o,S_o}]$ across all four prompts. 
For occupations that skew male in the training data, bias is amplified if it skews further male in generated images, and vice versa for occupations that skew female. 
Bias decreasing from training to generation is considered de-amplification. 
We exclude occupations that exhibit different directions of bias at training and generation from our analysis.

\begin{table*}[t]
 \small
  \centering 
  \begin{tabular}{p{0.45\linewidth} p{0.45\linewidth}}
    \toprule  
    \bf Caption & \bf Details \\
    \midrule
    Portrait of smiling young female \textbf{mechanic} inspecting a CV joint on a car in an auto repair shop & Contains person description (smiling young female), activity, and location \\ \addlinespace
    Muscular bearded \textbf{athlete} drinks water after good workout session in city park & Contains person description (muscular bearded), clues about attire (workout clothes), and activity \\ \addlinespace
    Portrait of a \textbf{salesperson} standing in front of electrical wire spool with arms crossed in hardware store & Contains activity, information about surroundings, and location \\
    \bottomrule
  \end{tabular}
  \caption{Training captions often include additional context and details (e.g., descriptions, activity information) that reduce ambiguity. As shown in these examples, captions may contain explicit and implicit gender information. In contrast, the prompts we use to generate images (Table \ref{tab:prompts}), lack specificity and convey limited information.}
  \vspace{-1.0em}
  \label{tab:cap_examps}
\end{table*}

\section{Naive Approach} \label{sec:keyword-querying}
We examine the extent to which Stable Diffusion amplifies gender-occupation biases from the data by selecting training examples that contain a given occupation in the caption (e.g., all captions that contain the word ``president'').
In practice, we randomly sample a subset of 500 training examples as opposed to using all examples.
We find that Stable Diffusion amplifies bias relative to the training data by 12.57\%\footnote{We report values for Stable Diffusion 1.4 throughout the paper, but results for both model versions are presented in Table \ref{tab:results}. Overall, we observe similar trends for both models.} on average across all occupations and prompts (10.24\% for Prompt \#1, as shown in Figure \ref{fig:base}). 
This behavior is concerning because instead of reflecting the training data and its statistics, the model compounds bias by further underrepresenting groups. 
However, when qualitatively inspecting examples, we observe discrepancies in how occupations are presented in captions vs. prompts due to varying levels of ambiguity.

For example, we notice the use of explicit \textit{gender indicators} to emphasize deviations from stereotypical gender-occupation associations, such as female mechanics.
While gender information is used frequently in captions, we hypothesize that usage is more common for underrepresented groups. 
If this hypothesis holds, the gender distribution would shift closer towards balanced in resulting training images.
As a result, the decision to focus on all captions vs. captions without any gender indicators might exaggerate amplification measures.

More generally, prompts commonly used to study gender-occupation bias are intentionally underspecified, or lack detail. 
Underspecification results in the model having to generate images from textual inputs that are vague and open to interpretation \citep{hutchinson-etal-2022-underspecification, mehrabi-etal-2023-resolving}. 
For instance, the prompt ``A photo of the face of a/an [OCCUPATION]'' does not contain any adjectives or information about surroundings, activities, etc.
In contrast, captions may contain context and details that result in less ambiguous descriptions, as shown in Table \ref{tab:cap_examps}.\footnote{We showcase examples that include descriptions of individuals and activities they are engaged in.}

\definecolor{salmon}{HTML}{E8ADAA}
\definecolor{lavender}{HTML}{CCCCFF}
\begin{figure}[t]
    \centering
    \includegraphics[width=0.7\linewidth]{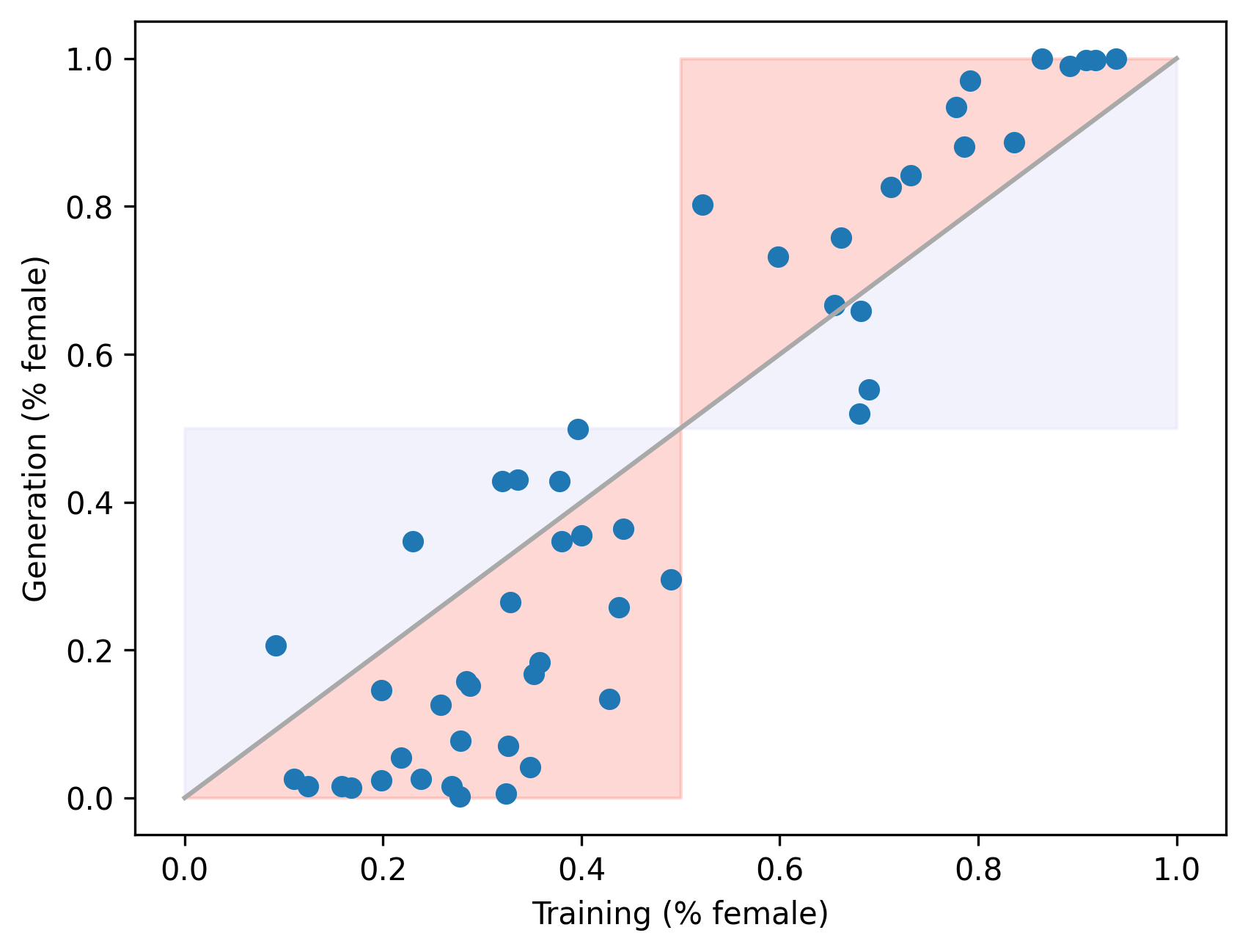}
    \caption{{\bf Bias amplification for our initial naive approach.} There appears to be consistent bias amplification between training and generation. The x-axis corresponds to the \% female in training images, and the y-axis corresponds to the \% female in generated images (using Prompt \#1). Each point represents an occupation. Shading: \textbf{\textcolor{salmon}{Amplification}} and \textbf{\textcolor{lavender}{De-Amplification}}.}
    \vspace{-1.0em}
    \label{fig:base}
\end{figure}

\begin{figure*}
     \centering
     \begin{subfigure}[b]{0.47\textwidth}
         \centering
         \includegraphics[width=\textwidth]{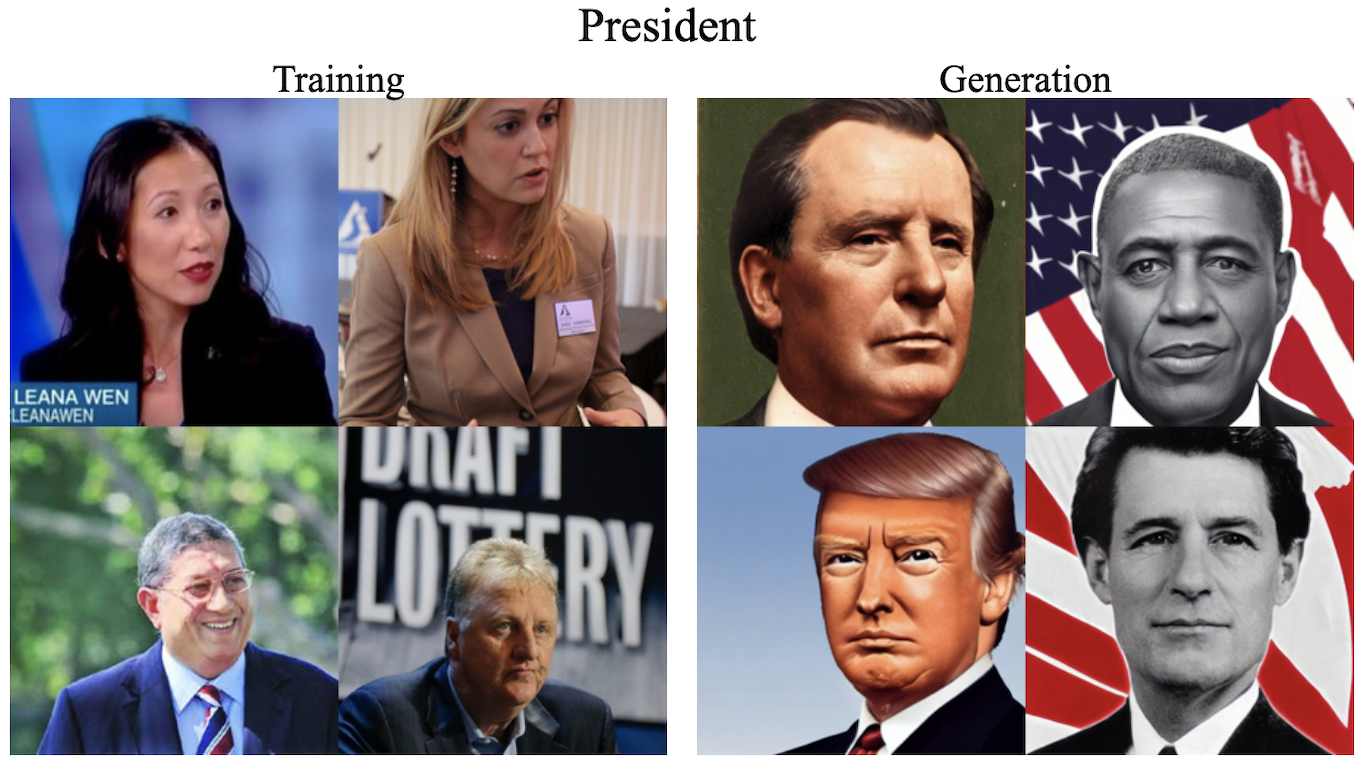}
         \caption{Training captions for \textbf{President}: 1) "Leana Wen, Planned Parenthood president..." 2) "New Schaumburg Business Association President..." 3) "BCCI president N Srinivasan..." 4) "Indiana Pacers president of basketball operations..."}
         \label{fig:pres}
     \end{subfigure}
     \hfill
     \begin{subfigure}[b]{0.47\textwidth}
         \centering
         \includegraphics[width=\textwidth]{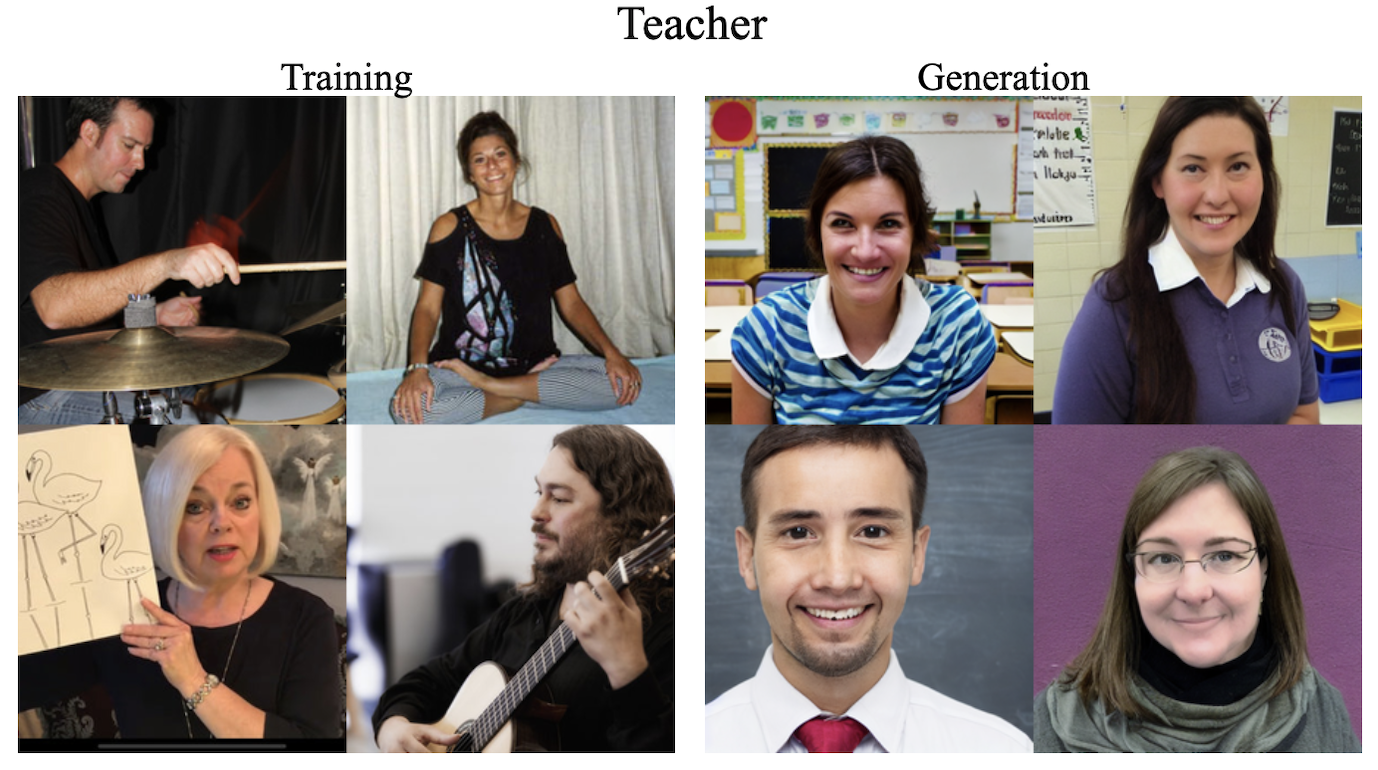}
         \caption{Training captions for \textbf{Teacher}: 1) "Brad Draper, percussion teacher..." 2) "teacher/author in the 80s sits in yoga lotus pose..." 3) "Jo Anne Young Art Teacher..." 4) "patrick oconnell Classical Guitar Teacher..."}
         \label{fig:teach}
     \end{subfigure}
        \caption{We handpick examples of discrepancies in how occupations are depicted in training vs. generated examples for \textbf{President} (left) and \textbf{Teacher} (right) professions using our naive approach.}
        \vspace{-1.0em}
        \label{fig:pres_teach}
\end{figure*}

Discrepancies in how captions and prompts are written also impact how occupations are depicted in training and generated images. 
These differences are especially notable for occupations that have multiple interpretations. 
For example, when querying for training examples containing ``president'', the resulting captions may refer to various types of presidents, including the president of a company or organization, as shown in Figure \ref{fig:pres}.
However, when generating images using the prompt ``A photo of the face of a president'', the model appears to interpret president as a leader of a country, often the United States (we also showcase similar differences for the occupation teacher in Figure \ref{fig:teach}). 

To make reasonable comparisons between bias at training vs. generation, we should compare gender ratios over similar captions and prompts. 
Therefore, we cannot conclude whether differences in gender ratios between training and generation are due solely to the model amplifying bias, or other confounding factors that contribute to amplification. 
Next, we focus on reducing the impact of distribution shifts on bias amplification evaluation.

\section{Reducing Discrepancies}\label{sec:reducing-discrepancies}
In this section, we evaluate approaches to reduce training and generation discrepancies by restricting the search space of training examples. 
Note that prompts $P_o$ remain fixed, while the subset of training examples $S_o$ varies in our analysis.

\subsection{Captions Without Explicit Gender Indicators}
A notable distinction between training and generation is the use of explicit gender indicators, which is absent from prompts. 
On average, more than half the captions (59.5\%) contain explicit gender information.
Furthermore, gender usage in captions varies depending on which gender is underrepresented for a given occupation.
For example, images of female mechanics in the training data frequently accompany captions that indicate the mechanic is female.
In comparison, this specification is less common for male mechanics (only 30\% of male mechanic examples contain explicit gender indicators, as opposed to 68\% for female mechanics). 

To validate these observations, we compute the correlation between the percentage of females in training images and the percentage of captions with female indicators.
We expect that female-skewing occupations are less likely to contain explicit female gender indicators in captions, resulting in a negative correlation.
The Pearson's correlation coefficient is indeed negative, with a coefficient value of -0.458 and statistically significant (significance level $<0.05$).
These results suggest that including training examples with gender information during evaluation may exaggerate amplification.

\paragraph{Addressing Gender Indicators}

To assess whether amplification differs for the subset of captions without indicators, we split the training examples selected in Section \ref{sec:keyword-querying} by detecting direct gender mentions in the captions (more details in \ref{sec:gender-indicacors}).
We focus on the subset of captions, $S_o$, without explicit male or female indicators.

\paragraph{Reduced Bias Amplification}
We observe that bias amplification is noticeably lower when focusing on the no-gender indicator subset of training examples. 
Compared to the initial amplification of 12.57\% for keyword querying, the average amplification for captions without gender indicators is 8.66\% ($\downarrow$ 31\%), as shown in Table \ref{tab:results}.
This behavior aligns with the reasoning described above --- gender indicators are more likely to delineate the presence of the underrepresented gender, which in turn inflates amplification measures. 

\subsection{Nearest Neighbors (\textsc{NN})}
Beyond explicit gender indicators, there are clear differences in the information conveyed by prompts vs. captions.
The prompts we use are concise and structured, but lack concrete details.
On the other hand, randomly sampled training captions are more diverse and vary in their usage of the occupation and the amount of contextual information, as highlighted in Table \ref{tab:cap_examps} and Figure \ref{fig:pres_teach}.
Furthermore, captions may contain implicit gender information (e.g., descriptors, attire, activity details) which is absent from prompts.

These qualitative differences are also apparent when comparing caption and prompt text embeddings. 
We use Sentence-BERT \citep{reimers-2019-sentence-bert} to compute text embeddings,\footnote{We use the all-MiniLM-L6-v2 model for text embeddings.} and calculate the average pairwise cosine similarity between caption and prompt embeddings for each occupation.
We find that the average cosine similarity across occupations is 0.385, indicating that captions and prompts are highly dissimilar (relative to nearest neighbors, which we will see next).

\begin{figure}[t]
     \centering
     \begin{subfigure}[t]{0.48\linewidth}
         \centering
         \includegraphics[width=\linewidth]{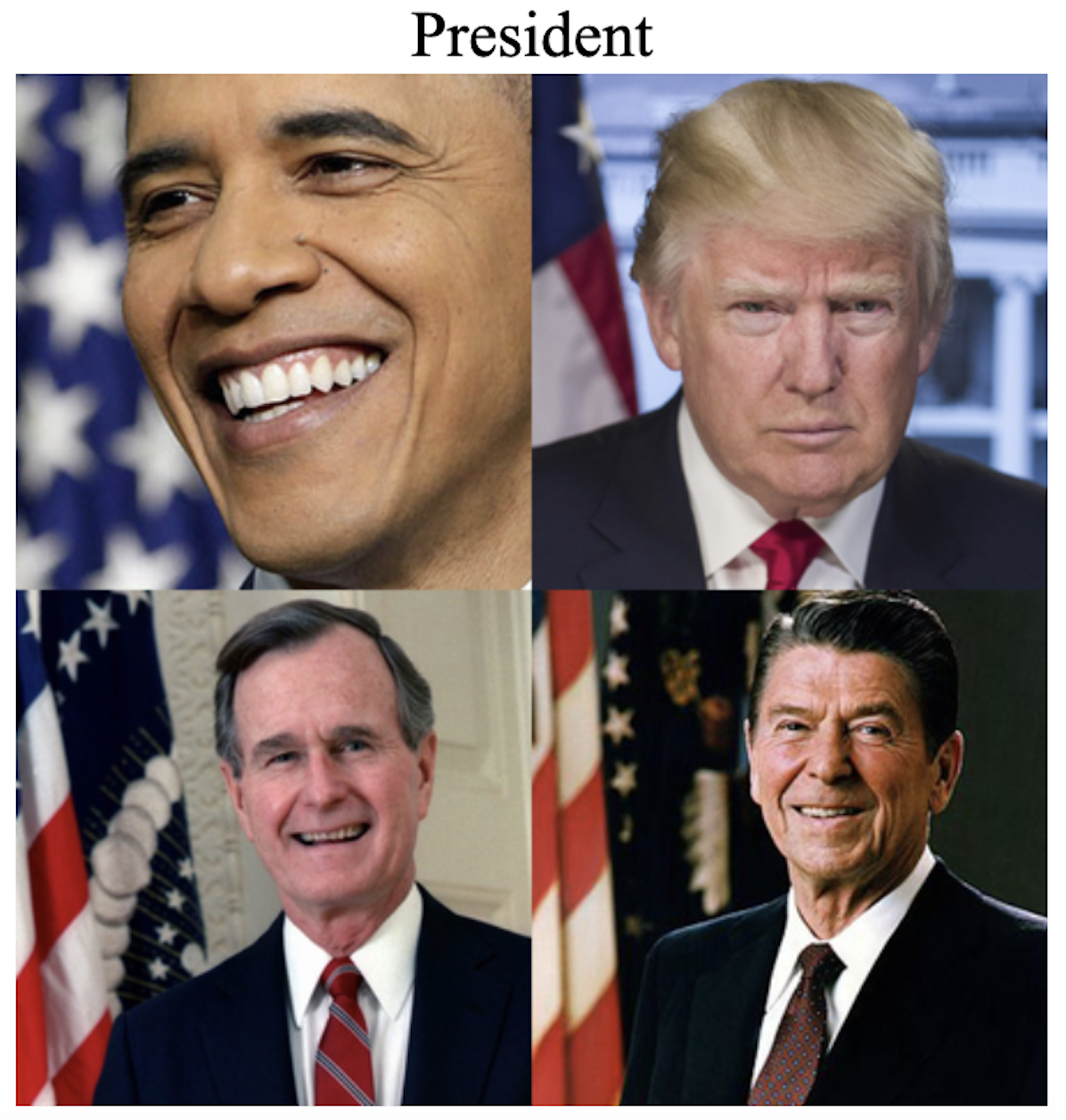}
         \caption{Training captions for \textbf{President}: 1) "The president is pictured smiling." 2) "President Donald J. Trump - Official Photo" 3) "Portrait of President George H. W. Bush" 4) "Official Portrait of President Ronald Reagan"}
         \label{pres_nn}
     \end{subfigure}
     \hfill
     \begin{subfigure}[t]{0.48\linewidth}
         \centering
         \includegraphics[width=\linewidth]{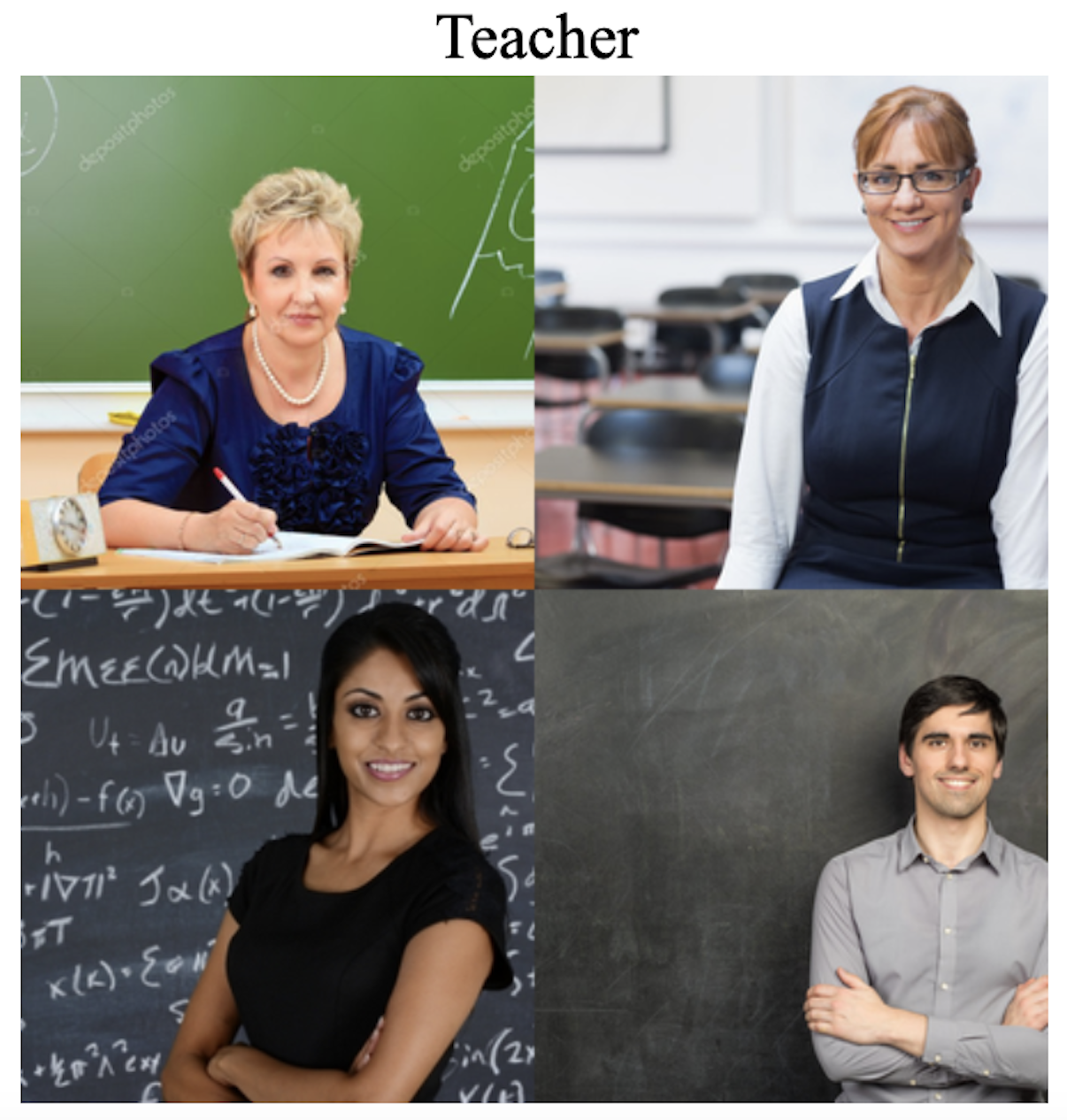}
         \caption{Training captions for \textbf{Teacher}: 1) "Picture of a teacher in the classroom" 2) "Portrait of a smiling teacher in a classroom." 3) "Portrait of teacher woman working" 4) "Teacher smiling in classroom, portrait"}
         \label{teach_nn}
     \end{subfigure}
        \caption{\textbf{Training examples chosen with Nearest Neighbors}. Selected training captions and images are more similar to prompts and generated images.} 
        \vspace{-1.0em}
        \label{fig:pres_teach_nn}
\end{figure}

\begin{table*}[t]
 \centering
 \small
  \begin{tabular}{l l cccc c}
    \toprule        
    \bf Approach & \bf Model & \bf Prompt \#1 & \bf Prompt \#2 & \bf Prompt \#3 & \bf Prompt \#4 & \bf Average \\
    \midrule
    \multirow{ 2}{*}{Naive Approach} & SD 1.4 & 10.24 & 17.57 & 10.77 & 11.68  & 12.57\\
   &  SD 1.5 & 10.87 & 16.36 & 11.15 & 9.91 & 12.07 \\
   \addlinespace
   \multirow{ 2}{*}{No Gender Indicators} & SD 1.4 & 6.49 & 13.58 & 7.09 & 7.49  & 8.66 \\
    & SD 1.5 & 6.76 & 12.41 & 6.82 & 5.87  & 7.97  \\
   \addlinespace
    \multirow{ 2}{*}{Nearest Neighbors} & SD 1.4 & 3.59 & 12.62 & 5.58 & 5.27  & 6.76 \\
    & SD 1.5 & 4.01 & 11.14 & 5.21 & 3.65 & 6.01  \\
   \addlinespace
   Nearest Neighbors + & SD 1.4 & 1.11 & 8.72 & 3.06 & 4.05   & 4.35 \\
   No Gender Indicators & SD 1.5 & 1.55 & 7.29 &2.78 & 2.72   & 3.59 \\
    \bottomrule
  \end{tabular}
   \caption{Bias Amplification across occupations using Stable Diffuson (SD) 1.4 and 1.5, for each prompt and averaged across prompts. Amplification lowers considerably when using nearest neighbors to select training captions and excluding captions with gender indicators. We see further reductions when combining approaches.}
  \label{tab:results}
  \vspace{-1.0em}
\end{table*}

\paragraph{Addressing Similarity Discrepancies}
To account for these gaps, we propose using nearest neighbors (\textsc{NN}) to select captions that closely resemble prompts.
We can find \textsc{NN} by considering all captions that contain a given occupation, and selecting examples based on the similarity between caption and prompt text embeddings instead of sampling randomly. 
As a result, the chosen captions are closer in structure and wording to prompts.
We use Sentence-BERT to obtain text embeddings and compute the cosine similarity between embeddings to measure the similarity between captions and prompts.\footnote{We acknowledge that the text embedding used for computing \textsc{NN} can reinforce certain biases. While perhaps CLIP and Sentence-BERT exhibit similar biases, our rationale for choosing the latter is to avoid leaking biases from Stable Diffusion's text encoder when selecting training examples.}
For a given occupation, we consider the top-$k$ similar captions, where $k=500$. 

Applying \textsc{NN}, the average cosine similarity between caption and prompt embeddings increases to 0.704 ($\uparrow$ 83\% from keyword querying), which occurs by design since we directly target examples that resemble prompts.
Note however, that the increase in similarity is also reflected in image embeddings.
The pairwise similarity of CLIP image embeddings increases with \textsc{NN} ($\uparrow$ 13\% from keyword querying), indicating that chosen training and generated images are slightly more similar. 

There are noticeable qualitative improvements as well. 
\textsc{NN} chooses captions that are closer in structure and meaning to prompts (e.g., ``Picture of a teacher in the classroom''), which also impacts corresponding training images.
As shown in Figure \ref{fig:pres_teach_nn}, the training images corresponding to \textsc{NN} captions for the word ``president'' primarily represent world leaders (often US presidents), while captions for the word ``teacher'' depict educators in classroom settings.\footnote{This behavior contrasts examples from Figure \ref{fig:pres_teach}, which showed various types of presidents and teachers.}

\paragraph{Reduced Bias Amplification}
When selecting training examples $S_o$ using \textsc{NN}, we see that bias amplification reduces considerably across occupations and prompts, as shown in Table \ref{tab:results}.
The average amplification drops to 6.76\% ($\downarrow$ 46\% relative to keyword querying).
While \textsc{NN} yields increased similarity between training and generated examples, there are still unresolved sources of distribution shift that impact amplification measures.

\subsection{Combining Approaches}
\begin{figure*}[ht]
     \centering
     \begin{subfigure}[t]{0.31\textwidth}
     \captionsetup{justification=centering}
         \centering
         \includegraphics[width=\textwidth]{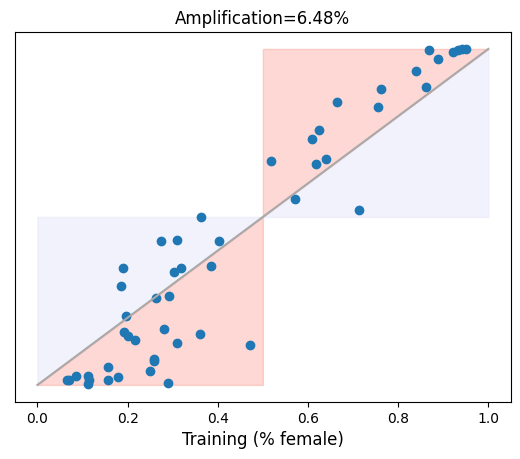}
         \caption{Captions without Gender Indicators}
         \label{fig:combined2}
     \end{subfigure}
     \hfill
     \begin{subfigure}[t]{0.34\textwidth}
     \captionsetup{justification=centering}
         \centering
         \includegraphics[width=\textwidth]{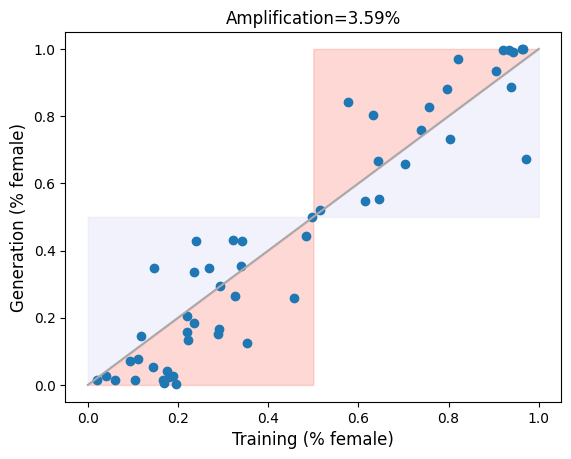}
         \caption{Nearest Neighbors}
         \label{fig:combined1}
     \end{subfigure}
     \hfill
     \begin{subfigure}[t]{0.31\textwidth}
     \captionsetup{justification=centering}
         \centering
         \includegraphics[width=\textwidth]{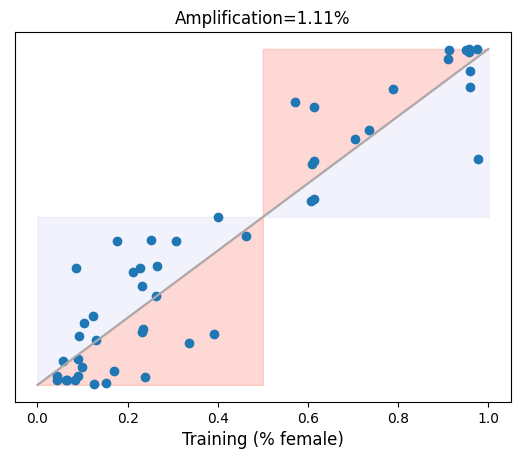}
         \caption{Combined Approach}
         \label{fig:combined3}
     \end{subfigure}
        \caption{{\bf Bias amplification for various approaches to address discrepancies between training and generation.} The proposed approaches yield lower bias amplification, especially the combined method (c). Results are shown for Prompt \#1. Regions are shaded based on \textbf{\textcolor{salmon}{Amplification}} and \textbf{\textcolor{lavender}{De-Amplification}}.}
        \label{fig:combined}
        \vspace{-1.0em}
\end{figure*}

We observe that amplification further reduces when combining the no-gender indicator subset with \textsc{NN}, as shown in the last rows in Table \ref{tab:results}. 
The average amplification decreases to 4.35\%, which is noticeably lower compared to the values for each method individually. 
Both methods work in tandem to reduce distributional differences in complementary ways, perhaps by targeting both explicit and implicit gender information. 
We also observe greater reductions for specific prompts; as shown in Figure \ref{fig:combined3}, amplification is just 1.11\% for Prompt \#1.

We perform a one-sample t-test to test the null hypothesis that the expected amplification is 0 for each of the prompts; we fail to reject the null hypothesis for prompts \#1 and \#3 and reject the null hypothesis for prompts \#2 and \#4  (significance level $<0.05$).
Our results indicate a portion of amplification is unexplained for all prompts, especially prompts \#2 and \#4, and may involve more subtle confounding factors. 
Although the proposed methods do not account for all possible discrepancies between training and generation, we observe that the bias measures become closer as we select subsets of training captions that resemble prompts.

\section{Removing Distributional Differences: A Lower Bound} \label{sec:captions-as-prompts}

While the previous approaches reduce discrepancies between training and generation by evaluating amplification with captions that are more similar to prompts, we can instead modify the prompts we use to align with captions more closely.
In this setup, we use the original training subset ($S_o$) from Section \ref{sec:keyword-querying}, but the prompts ($P_o$) now match the captions verbatim. 
For every prompt in $P_o$, we generate 10 images, and then compute amplification using $P_o:=S_o$ for each occupation.

\begin{figure}
     \centering
     \begin{subfigure}[t]{0.49\linewidth}
     \captionsetup{justification=centering}
         \centering
         \includegraphics[width=\linewidth]{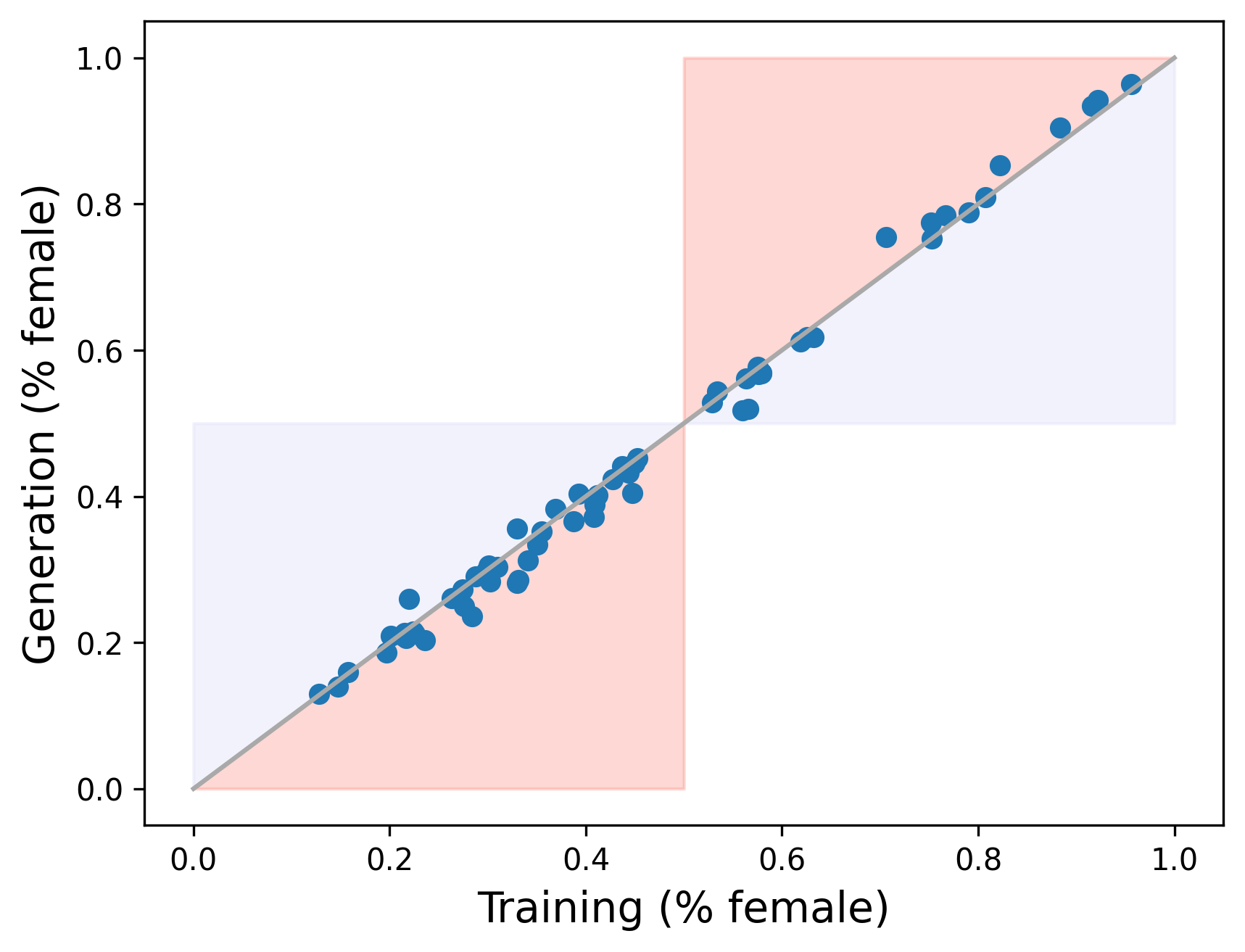}
         \caption{All Captions}
         \label{fig:exact1}
     \end{subfigure}
     \hfill
     \begin{subfigure}[t]{0.45\linewidth}
     \captionsetup{justification=centering}
         \centering
         \includegraphics[width=\linewidth]{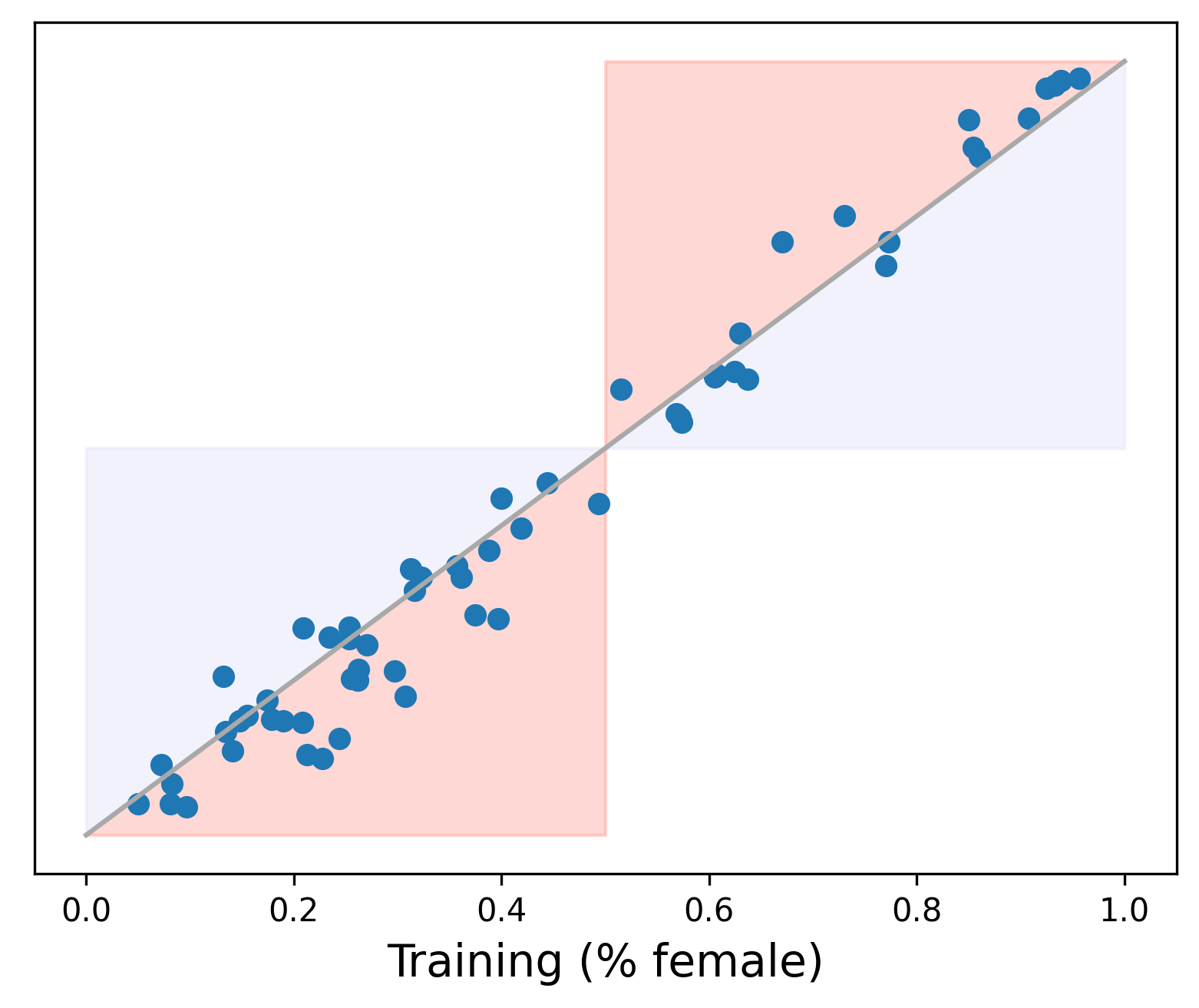}
         \caption{Captions without Gender Indicators}
         \label{fig:exact2}
     \end{subfigure}
        \caption{{\bf Bias amplification when prompting with training captions.} We observe minimal amplification when $P_o=S_o$ (left). This behavior mostly holds when focusing on captions without explicit gender indicators (right). Shading: \textbf{\textcolor{salmon}{Amplification}} and \textbf{\textcolor{lavender}{De-Amplification}}.}
        \vspace{-1.0em}
        \label{fig:exact}
        
\end{figure}

By design, this approach removes mismatches between captions and prompts, since prompts and captions are now identical.
We then ask: \textit{Does using identical texts to measure training and generation bias lower amplification?} 
We hypothesize that enforcing prompts and captions to match yields similar bias measurements, which in turn reduces amplification. 
As shown in Figure \ref{fig:exact1}, amplification is minimal when $P_o=S_o$ and most occupations reside along the diagonal (no amplification). 
The average amplification drops to 0.68\%, indicating that the model mostly reflects training bias.\footnote{\label{sig}However, we reject the null hypothesis that the expected amplification is 0 using a one-sample t-test.}
Furthermore, amplification remains consistently low, even for occupations that are highly imbalanced.

For captions that contain either male or female gender indicators, the model generates images that match the gender of corresponding training images (with 98.41\% accuracy), since this information is directly provided in the prompt. 
Therefore, we analyze results separately on the subset of captions without gender indicators. 
As shown in Figure \ref{fig:exact2}, bias amplification is larger for the no gender indicator subset as compared to all captions. 
That being said, the average amplification remains low at 2.05\% ($\downarrow$ 84\% relative to keyword querying).\footref{sig}

Although practitioners are unlikely to utilize prompts that exactly match training captions (nor do we make this recommendation), this experiment highlights the impact of distributional similarity between captions and prompts when comparing biases. 
In addition, it provides a lower bound to the bias amplification problem.
In summary, we conclude that the model nearly mimics biases from the data when prompted with training captions.

\definecolor{yellowgold}{HTML}{FFCE44}
\definecolor{limegreen}{HTML}{6CC417}
\begin{figure*}
     \centering
     \begin{subfigure}[b]{0.22\textwidth}
     \captionsetup{justification=centering}
         \centering
         \includegraphics[width=\textwidth]{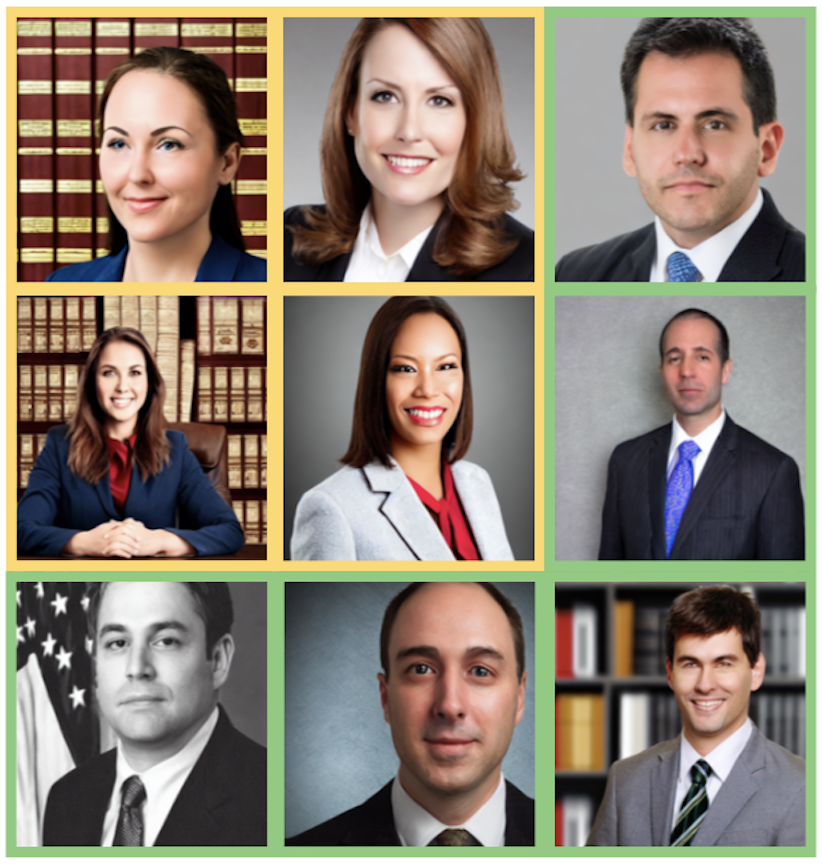}
         \caption{A photo of the face of an attorney}
         \label{fig:attorney1}
     \end{subfigure}
     \hfill
     \begin{subfigure}[b]{0.22\textwidth}
     \captionsetup{justification=centering}
         \centering
         \includegraphics[width=\textwidth]{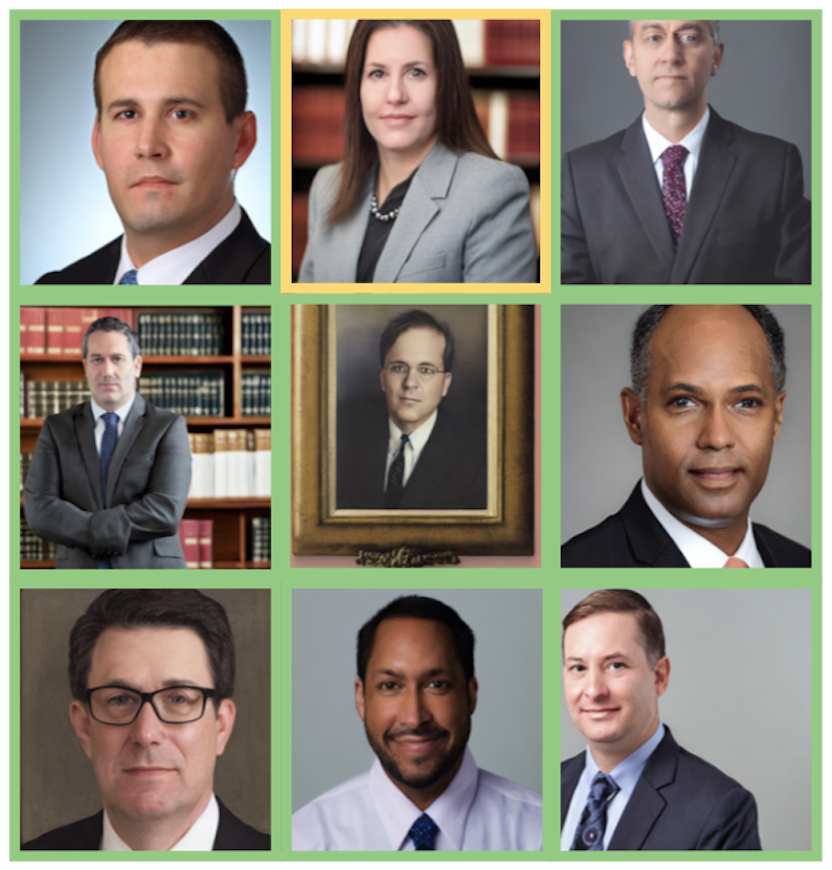}
         \caption{A portrait photo of an attorney}
         \label{fig:attorney2}
     \end{subfigure}
     \hfill
     \begin{subfigure}[b]{0.22\textwidth}
     \captionsetup{justification=centering}
         \centering
         \includegraphics[width=\textwidth]{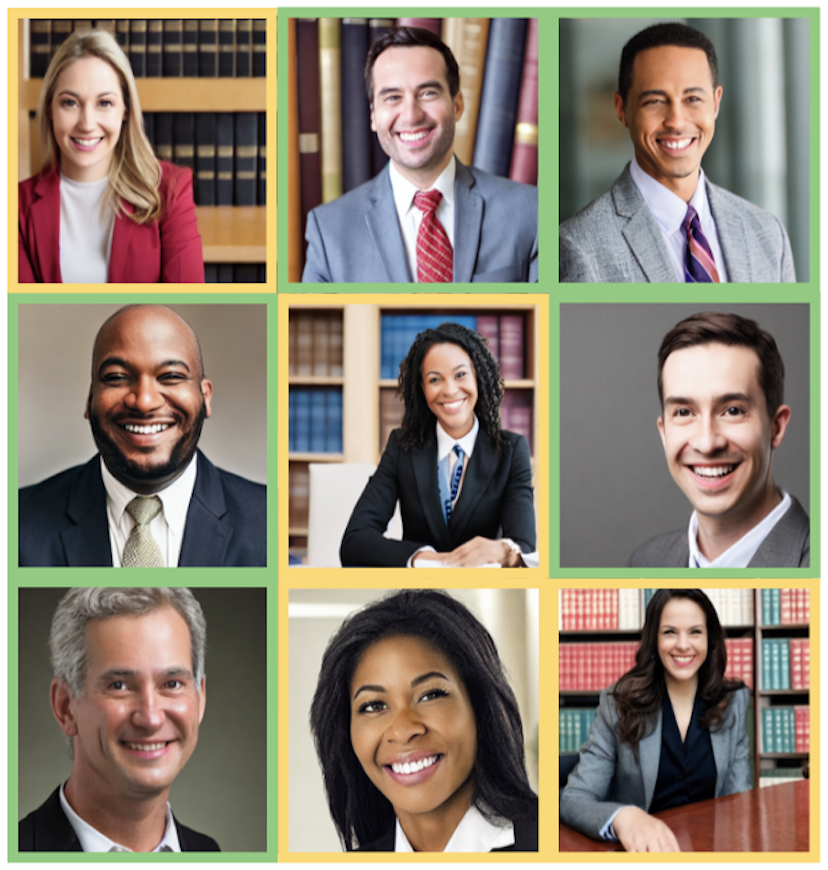}
         \caption{A photo of an attorney smiling}
         \label{fig:attorney3}
     \end{subfigure}
     \hfill
     \begin{subfigure}[b]{0.22\textwidth}
     \captionsetup{justification=centering}
         \centering
         \includegraphics[width=\textwidth]{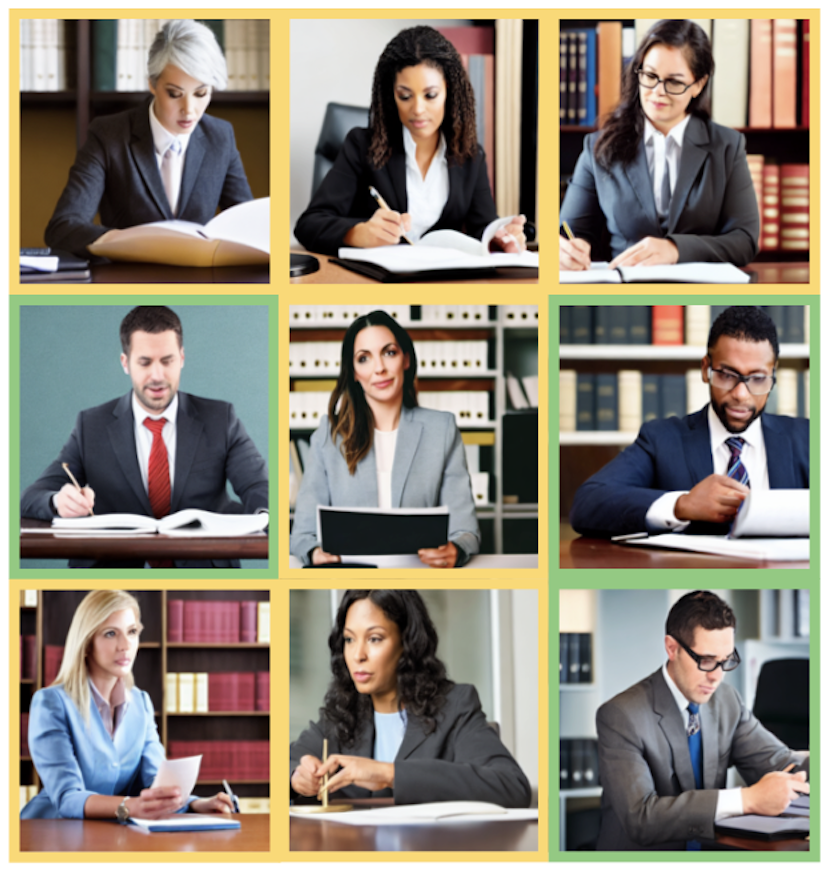}
         \caption{A photo of an attorney at work}
         \label{fig:attorney4}
    \end{subfigure}
        \caption{\textbf{Generated examples for the occupation attorney using different prompts.} Specific wording and phrasing choices in prompts lead to noticeable differences in the \% of generated images that are female (boxes are colored based on \textbf{\textcolor{yellowgold!90!black}{predicted female}} and \textbf{\textcolor{limegreen!90!black}{predicted male}}). Although we only include 9 images per prompt here, these proportions are similar to what is exhibited in the 500 generated images.}
        \label{fig:var_prompt}
        \vspace{-1.0em}
\end{figure*}

\section{Related Work}

\paragraph{Relating pretraining data to model behavior} 
There is a growing body of work focused on studying pretraining data properties and statistics, as well as understanding their relation to model behavior. 
This type of large-scale data and model analysis provides useful insights into model learning and generalization capabilities \citep{carlini2023extracting}. 
Recent work shows that few-shot capabilities of large language models are highly correlated with pretraining term frequencies, and that models struggle to learn long-tail knowledge \citep{kandpal2022large,razeghi-etal-2022-impact}. 
Several works have also explored the relationship between pretraining data and model performance from a causal perspective \citep{biderman2023pythia, elazar2023measuring, longpre2023pretrainers}.
For example, \citet{longpre2023pretrainers} comprehensively investigate how various data curation choices and pretraining data slices affect downstream task performance. 


\paragraph{Bias Amplification} 
Our work is strongly inspired by the findings of \citet{zhao-etal-2017-men}, who show that structured prediction models amplify biases present in the data. 
However, there are important differences to note.
First, their task involves jointly predicting multiple target labels, including gender, as opposed to generating images.
Additionally, their work focuses on mitigating amplification, as opposed to investigating various factors that contribute to amplification.
\citet{hall-et-al2022} consider how data, training, and model-related choices influence amplification using a classification setup with synthetic bias, but do not examine how distribution shifts contribute to amplification.

\citet{friedrich2023FairDiffusion} also compare biases exhibited by LAION and Stable Diffusion, and show that the model displays bias amplification. 
Instead of identifying relevant training examples using captions, they use text-image similarity between prompts and training images. 
Similar to \citet{zhao-etal-2017-men}, their work mainly focuses on mitigating model bias using their proposed method, fair diffusion, while our work is centered around analyzing confounding factors that impact amplification. 

\paragraph{Bias in text-to-image models}
While it is well-established that language and vision models are susceptible to biases individually, recent work has shown that text-to-image models are prone to similar biases. 
Several works analyze various biases in text-to-image models, including underexplored topics such as geographical disparities 
\citep{Basu_2023_ICCV, naik2023social} and intersectional biases \citep{fraser2023friendly, luccioni2023stable}.
\citet{bianchi-et-al2022} demonstrate that stereotypes persist even after prompting the model with counter-stereotypes.
However, these works primarily focus on evaluating model biases, and do not examine the training data. 

\section{Discussion}\label{sec:discussion}

\paragraph{Generalizability} Our work demonstrates that using naive procedures to evaluate bias amplification can lead to exaggerated amplification measures.
While our analysis does not account for all sources of distribution shift that contribute to amplification, it is meant to be illustrative.
We encourage future studies to build on our findings by examining different experimental setups (i.e., datasets, models, and types of bias) to gain a more comprehensive understanding of bias amplification and the impact of confounding factors.

\paragraph{Variation Across Prompts} As we highlight in Figure \ref{fig:var_prompt}, small changes to prompts can have a resounding effect on conclusions about model bias. 
For example, “A portrait photo of an attorney” skews heavily male while “A photo of an attorney at work” skews female in generated images. 
Furthermore, reductions in amplification differ based on the prompt (e.g., Prompt \#1 exhibits an 89\% reduction as opposed to 49\% for Prompt 2),  indicating that the confounding factors have a varying impact.
These results suggest that there may also be prompt-specific sources of distribution shift, which is an important consideration when choosing prompts.

\paragraph{Amplification Baseline} Our interpretation of bias amplification is centered around models exacerbating biases found in the training data as opposed to real-world statistics \citep{Kirk2021BiasOA, bianchi-et-al2022}.
Both approaches are useful to study but answer fundamentally different questions.
Our approach offers insights into whether model behavior reflects the training data, while real-world amplification captures how well the data and model together reflect reality.

\paragraph{Connection to Simpson's Paradox} The title of our paper alludes to Simpson's Paradox \citep{simpsons-paradox}, a phenomenon in which a trend or relationship observed in subgroups within the data reverses or disappears when subgroups are combined. 
We draw direct parallels to our analysis and insights; although we observe substantial amplification in our initial setup, amplification reduces drastically after selecting specific subsets of the training data and decreasing the impact of confounding factors.

\paragraph{Recommendations}
Our findings underscore how distribution shifts contribute to bias amplification, which has important implications.
Those involved in data-focused efforts should consider how practitioners specify prompts and interact with models when curating training data.
Alternatively, crowdsourcing or automatically rewriting existing training captions to reflect real-world model usage may result in lower amplification.
Additionally, we recommend that evaluations use multiple prompts and remove prompt-specific confounding factors (e.g., by using \textsc{NN} to select relevant training examples).

\section{Conclusion}
In summary, we investigate whether Stable Diffusion amplifies gender-occupation biases by comparing training data and model biases. 
We highlight how naive evaluations of amplification fail to consider distributional differences between training and generation, which leads to a misleading understanding of model behavior.
Although amplification is not eliminated entirely, we observe that reducing discrepancies between captions and prompts during evaluation results in substantially lower measurements.
We strongly recommend that any analysis comparing training data and model biases, or any dataset and model properties more generally, account for various distribution shifts that skew evaluations.

\section*{Limitations}
Beyond the training data, another source of bias is the text embeddings obtained from CLIP.
By solely comparing biases in the data vs. those exhibited by Stable Diffusion, our analysis overlooks biases that arise from encoding prompts. 
As a result, we cannot disentangle how much this component impacts overall amplification. 
Note that the effect of such an external embedding cannot be easily accounted for, since CLIP's training data is not public.
More work is needed to understand the impact of using external, frozen models as a model component.

Additionally, we automate gender classification using CLIP because previous works have shown that CLIP gender predictions align with human annotations and CLIP gender classification performance on the FairFace dataset\footnote{https://github.com/joojs/fairface} is strong ($>95\%$) across various racial categories. 
Nevertheless, we recognize the limitations of using a model to classify gender in images, since CLIP inherits biases from its training data.

\section*{Ethics Statement}
\paragraph{Scope of Work} Our work centers around critically examining bias amplification evaluation.
The approaches we propose to reduce distribution shifts observed during evaluation do not solve underlying gaps between the data used to train models and how users interact with models.
Rather, they serve to deepen our understanding of why models amplify biases present in the training data. 
Ideally, our findings will motivate future work on 1) thorough and nuanced evaluations of bias amplification and 2) fundamentally addressing training and generation discrepancies from a data perspective. 

\paragraph{Bias Definition} Our work focuses on a narrow slice of social bias analysis by studying gender-occupation stereotypes. 
However, since models exhibit various types of discriminatory bias (e.g., racial, age, geographical,
socioeconomic, disability, etc.), as well as intersectional biases, it is equally important to perform evaluations for these definitions of bias. 
Furthermore, we only consider binary gender, which has clear drawbacks. Our analysis ignores how text-to-image models perpetuate biases for non-binary identities and relies on information such as appearance and facial features to infer gender in training and generated images, which can further propagate gender stereotypes.

\paragraph{Geographical Diversity} The captions and prompts used to study bias are solely written in English. 
We hope future work will shed light on multilingual bias amplification in text-to-image models. 
It is also worth noting that the gender-guesser library (infers gender from names) likely performs worse on non-Western names. 
The documentation mentions that the library supports over 40,000 names and covers a ``vast majority of first names in all European countries and in some overseas countries (e.g., China, India, Japan, USA)''. 
Therefore, the name coverage (or lack thereof) impacts our ability to identify captions with gender information.

\bibliography{anthology,custom}

\begin{table*}[ht]
  \centering 
  \begin{tabular}{p{0.18\linewidth} p{0.18\linewidth} p{0.18\linewidth} p{0.18\linewidth} p{0.18\linewidth}}
\toprule 
& & \bf Occupations & &\\
\midrule
accountant & dentist & journalist & poet & singer \\
architect & dietitian & lawyer & politician & student \\
assistant & doctor & librarian & president & supervisor \\
athlete & engineer & manager & prime minister & surgeon \\
attorney & entrepreneur & mechanic & professor & teacher \\
author & fashion designer & musician & programmer & technician \\
baker & filmmaker & nurse & psychologist & therapist \\
bartender & firefighter & nutritionist & receptionist & tutor \\
ceo & graphic designer & painter & reporter & veterinarian \\
chef & hairdresser & pharmacist & researcher & writer \\
comedian & housekeeper & photographer & salesperson &  \\
cook & intern & physician & scientist & \\
dancer & janitor & pilot & senator & \\
\bottomrule
\end{tabular}
  \caption{List of 62 occupations used to study gender-occupation biases.}
  \label{tab:occs}
\end{table*}

\begin{figure*}[ht]
     \centering
     \begin{subfigure}[t]{0.32\textwidth}
     \captionsetup{justification=centering}
         \centering
         \includegraphics[width=\textwidth]{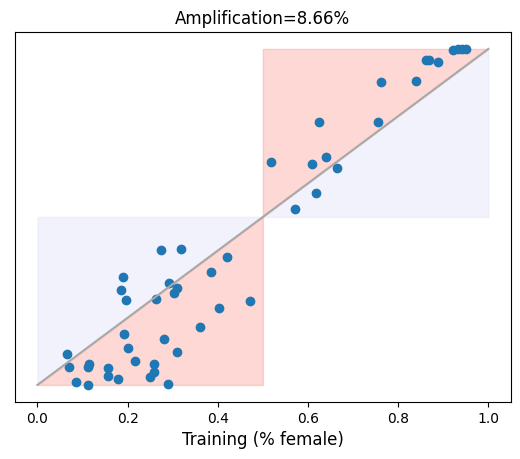}
         \caption{Captions without Gender Indicators}
         \label{fig:combined_avg2}
     \end{subfigure}
     \hfill
     \begin{subfigure}[t]{0.32\textwidth}
     \captionsetup{justification=centering}
         \centering
         \includegraphics[width=\textwidth]{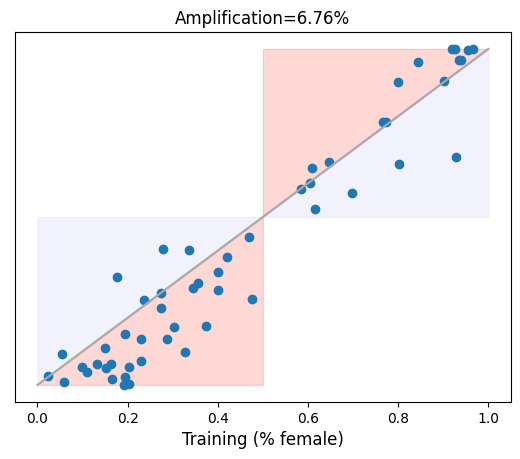}
         \caption{Nearest Neighbors}
         \label{fig:combined_avg1}
     \end{subfigure}
     \hfill
     \begin{subfigure}[t]{0.32\textwidth}
     \captionsetup{justification=centering}
         \centering
         \includegraphics[width=\textwidth]{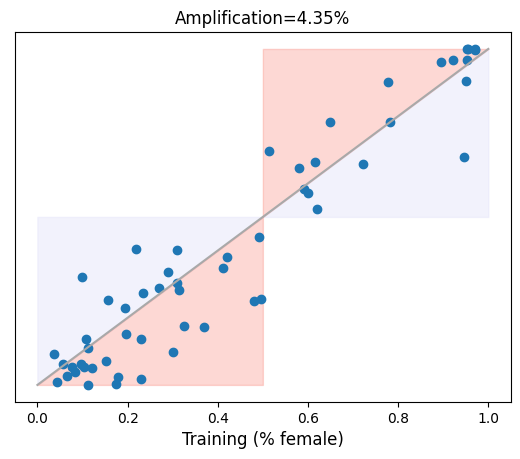}
         \caption{Combined Approach}
         \label{fig:combined_avg3}
     \end{subfigure}
        \caption{{\bf Bias amplification for various approaches to address discrepancies between training and generation.} The proposed approaches yield lower bias amplification, especially the combined method (c). Results are averaged across all prompts. Regions are shaded based on \textbf{\textcolor{salmon}{Amplification}} and \textbf{\textcolor{lavender}{De-Amplification}}.}
        \label{fig:combined_avg}
\end{figure*}

\newpage
\appendix
\section{Appendix}
\label{sec:appendix}

\subsection{List of Occupations}
A full list of occupations is shown in Table \ref{tab:occs}.
\\ \\
We exclude occupations that exhibit different directions of bias at training and generation from our amplification results, since this behavior does not adhere to our definition of amplification. 
There are 5 occupations (assistant, author, dentist, painter, supervisor) that exhibit switching behavior consistently for all prompts, using both SD 1.4 and 1.5.
More research is needed to understand and explain this behavior.

\subsection{Image Gender Classification} 
While CLIP is susceptible to biases \citep{hall2023visionlanguage}, its gender predictions have been shown to align with human-annotated gender labels \citep{bansal-et-al2022, Cho2022DallEval}. 
In addition, we perform human evaluation with 7 participants on 200 randomly selected training and generated images. 
We ask participants to provide binary gender annotations (or indicate that they are unsure), and find that Krippendorff's coefficient, which measures inter-annotator agreement, is high ($\alpha=0.948$). Additionally, 98\% of CLIP predictions match the majority vote annotations.

\subsection{Explicit Gender Indicators}
\label{sec:gender-indicacors}
To identify captions with explicit gender information, we consider 1) gender words (male, female, man, woman, gent, gentleman, lady, boy, girl), 2) binary gender pronouns (he, him, his, himself, she, her, hers, herself), and 3) names. 
We perform named entity recognition using the \textit{en\_core\_web\_lg} model from spaCy to identify name mentions, and then use the gender-guesser library \url{https://pypi.org/project/gender-guesser/} to infer gender. 

\begin{table*}
\centering
\small
\begin{tabular}{l|c|ccccc}
\toprule
      Occupation &  Training &  Prompt 1 &  Prompt 2 &  Prompt 3 &  Prompt 4 &  SD (Prompts) \\
\midrule
      accountant &      29.8 &      29.5 &       3.4 &      43.8 &      35.7 &          15.1 \\
       architect &      31.4 &       4.2 &       2.2 &       3.0 &       0.0 &           1.5 \\
       assistant &      44.6 &      67.1 &      56.3 &      71.9 &      75.6 &           7.3 \\
         athlete &      44.8 &      80.0 &      51.9 &      69.3 &      77.3 &          11.0 \\
        attorney &      29.2 &      42.8 &       9.4 &      43.1 &      65.1 &          19.9 \\
          author &      42.8 &      83.6 &      53.0 &      81.5 &      61.0 &          13.1 \\
           baker &      41.4 &      81.1 &      31.2 &      58.8 &      59.3 &          17.7 \\
       bartender &      36.8 &      16.8 &       2.6 &      12.9 &      22.9 &           7.4 \\
             ceo &      15.0 &       2.6 &       1.8 &       4.8 &      11.9 &           4.0 \\
            chef &      28.0 &       7.0 &       1.2 &       1.4 &       5.8 &           2.6 \\
        comedian &      21.8 &       2.4 &       0.0 &       3.6 &       1.0 &           1.4 \\
            cook &      35.0 &      34.7 &       8.6 &      49.4 &      69.3 &          22.2 \\
          dancer &      81.0 &      88.7 &      98.8 &      99.0 &     100.0 &           4.6 \\
         dentist &      58.6 &      41.4 &       4.4 &      29.2 &      41.8 &          15.2 \\
       dietitian &      95.2 &     100.0 &     100.0 &     100.0 &      99.8 &           0.1 \\
          doctor &      40.8 &      33.7 &       3.8 &      14.6 &      57.6 &          20.5 \\
        engineer &      20.6 &       2.6 &       0.2 &       1.2 &       0.0 &           1.0 \\
    entrepreneur &      43.6 &      42.8 &       1.8 &      12.8 &      34.6 &          16.4 \\
fashion\_designer &      76.0 &      93.4 &      80.8 &      89.8 &      97.2 &           6.1 \\
       filmmaker &      29.2 &      12.6 &       3.2 &       8.3 &      14.9 &           4.5 \\
     firefighter &      14.6 &       1.6 &       1.0 &      15.9 &       3.2 &           6.1 \\
graphic\_designer &      52.8 &      11.8 &      14.4 &      32.7 &      41.6 &          12.5 \\
     hairdresser &      79.2 &      97.0 &      95.6 &      94.6 &      97.6 &           1.2 \\
     housekeeper &      91.4 &      99.0 &      99.8 &     100.0 &     100.0 &           0.4 \\
          intern &      57.6 &      65.8 &      31.5 &      77.2 &      53.4 &          17.0 \\
         janitor &      20.4 &       1.6 &       3.0 &      14.6 &       5.7 &           5.0 \\
      journalist &      38.4 &      49.9 &      59.9 &      68.8 &      64.0 &           7.0 \\
          lawyer &      27.6 &      26.5 &       8.0 &      39.0 &      47.7 &          14.9 \\
       librarian &      74.4 &      88.1 &      83.6 &      93.6 &      94.8 &           4.5 \\
         manager &      13.0 &      20.6 &       7.8 &      29.7 &      42.8 &          12.8 \\
        mechanic &      17.6 &       1.6 &       0.0 &       0.2 &      35.3 &          15.0 \\
        musician &      22.6 &       5.4 &       4.2 &       7.2 &       3.2 &           1.5 \\
           nurse &      88.8 &     100.0 &     100.0 &     100.0 &     100.0 &           0.0 \\
    nutritionist &      83.6 &      99.8 &      92.8 &      96.6 &      97.5 &           2.5 \\
         painter &      52.6 &      36.4 &      12.2 &      17.6 &       3.6 &          12.0 \\
      pharmacist &      68.0 &      84.2 &      26.9 &      54.9 &      91.7 &          25.6 \\
    photographer &      55.0 &      52.0 &      27.5 &      46.5 &      13.2 &          15.4 \\
       physician &      39.4 &      35.5 &       2.0 &      37.5 &      59.3 &          20.5 \\
           pilot &      30.4 &      34.7 &      12.2 &      66.3 &      15.9 &          21.4 \\
            poet &      30.8 &      15.2 &       2.0 &      19.5 &      32.8 &          11.0 \\
      politician &      21.6 &      14.5 &       4.2 &      15.9 &       9.6 &           4.6 \\
       president &      19.6 &       1.4 &       0.2 &       8.0 &       0.8 &           3.1 \\
  prime\_minister &      24.0 &      15.7 &      10.6 &      13.2 &      21.4 &           4.0 \\
       professor &      28.2 &       7.8 &       2.8 &       9.2 &       5.3 &           2.4 \\
      programmer &      23.0 &       0.2 &       0.0 &       0.2 &       0.0 &           0.1 \\
    psychologist &      58.6 &      44.3 &      21.6 &      57.2 &      52.9 &          13.8 \\
    receptionist &      91.4 &      99.8 &     100.0 &      99.8 &      99.8 &           0.1 \\
        reporter &      44.4 &      54.8 &      55.2 &      55.1 &      67.8 &           5.5 \\
      researcher &      44.6 &      80.2 &      41.8 &      67.6 &      50.9 &          14.8 \\
     salesperson &      39.8 &      43.0 &       5.2 &      33.1 &      33.7 &          14.2 \\
       scientist &      33.4 &      25.7 &      24.0 &      29.3 &      23.2 &           2.4 \\
         senator &      35.0 &      13.4 &       2.0 &       8.2 &       5.4 &           4.2 \\
          singer &      57.6 &      73.2 &      60.3 &      69.2 &      60.1 &           5.7 \\
         student &      63.0 &      55.3 &      48.5 &      62.1 &      43.3 &           7.1 \\
      supervisor &      65.2 &      18.3 &       4.8 &      16.6 &      14.9 &           5.2 \\
         surgeon &      30.2 &      82.5 &      15.6 &      67.6 &      82.5 &          27.5 \\
         teacher &      63.0 &      75.8 &      55.7 &      94.0 &      88.0 &          14.7 \\
      technician &      31.2 &       0.6 &       0.0 &       0.6 &       0.0 &           0.3 \\
       therapist &      74.8 &      82.6 &      63.3 &      79.2 &      87.5 &           9.0 \\
           tutor &      59.2 &      48.1 &      23.1 &      32.7 &      43.5 &           9.7 \\
    veterinarian &      55.2 &      66.7 &      44.7 &      64.1 &      89.9 &          16.0 \\
          writer &      30.2 &      73.3 &      30.1 &      76.0 &      63.8 &          18.3 \\
\bottomrule
\end{tabular}
\label{tab:occs_1.4}
\caption{The percentage of females across occupations in training images (using our initial approach from Section \ref{sec:keyword-querying}) and generated images using \textbf{SD 1.4}. We display generation results for each prompt, as well as the standard deviation (SD) across prompts.}
\end{table*}

\begin{table*}
\centering
\small
\begin{tabular}{l|c|ccccc}
\toprule
      Occupation &  Training &  Prompt 1 &  Prompt 2 &  Prompt 3 &  Prompt 4 &  SD (Prompts) \\
\midrule
      accountant &      29.8 &      34.9 &       5.4 &      42.1 &      45.2 &          15.8 \\
       architect &      31.4 &      10.0 &       2.2 &       2.2 &       3.4 &           3.2 \\
       assistant &      44.6 &      69.2 &      60.8 &      58.6 &      77.8 &           7.6 \\
         athlete &      44.8 &      76.6 &      46.0 &      50.0 &      74.3 &          13.8 \\
        attorney &      29.2 &      50.8 &      11.7 &      44.3 &      68.3 &          20.5 \\
          author &      42.8 &      88.2 &      57.4 &      75.4 &      69.0 &          11.1 \\
           baker &      41.4 &      82.3 &      33.9 &      53.3 &      66.6 &          17.7 \\
       bartender &      36.8 &      10.0 &       2.2 &       4.8 &      12.2 &           4.0 \\
             ceo &      15.0 &       1.4 &       2.0 &       5.4 &      18.5 &           6.9 \\
            chef &      28.0 &      12.0 &       0.8 &       1.4 &       7.0 &           4.6 \\
        comedian &      21.8 &       1.6 &       0.0 &       1.4 &       0.6 &           0.6 \\
            cook &      35.0 &      38.4 &      16.4 &      43.5 &      75.1 &          21.0 \\
          dancer &      81.0 &      83.8 &      97.4 &      97.6 &     100.0 &           6.4 \\
         dentist &      58.6 &      41.9 &       5.4 &      22.7 &      20.4 &          13.0 \\
       dietitian &      95.2 &     100.0 &     100.0 &     100.0 &      99.8 &           0.1 \\
          doctor &      40.8 &      38.2 &       8.8 &      12.6 &      53.4 &          18.4 \\
        engineer &      20.6 &      10.6 &       0.6 &       1.6 &       0.0 &           4.3 \\
    entrepreneur &      43.6 &      59.7 &       4.6 &      16.9 &      41.6 &          21.4 \\
fashion\_designer &      76.0 &      97.4 &      90.3 &      92.2 &      98.6 &           3.5 \\
       filmmaker &      29.2 &      18.4 &       5.2 &       8.8 &       7.8 &           5.0 \\
     firefighter &      14.6 &       1.4 &       0.2 &      12.5 &       4.5 &           4.8 \\
graphic\_designer &      52.8 &      22.6 &      15.3 &      29.5 &      63.3 &          18.4 \\
     hairdresser &      79.2 &      99.6 &      98.0 &      95.4 &      97.3 &           1.5 \\
     housekeeper &      91.4 &      99.6 &     100.0 &     100.0 &     100.0 &           0.2 \\
          intern &      57.6 &      72.6 &      37.1 &      68.8 &      60.4 &          13.8 \\
         janitor &      20.4 &       3.6 &       3.2 &       8.4 &       6.2 &           2.1 \\
      journalist &      38.4 &      57.2 &      60.2 &      59.7 &      60.7 &           1.4 \\
          lawyer &      27.6 &      34.1 &       8.8 &      36.8 &      48.2 &          14.4 \\
       librarian &      74.4 &      93.4 &      85.8 &      87.8 &      94.6 &           3.7 \\
         manager &      13.0 &      24.0 &      14.2 &      28.7 &      41.3 &           9.8 \\
        mechanic &      17.6 &       6.4 &       0.2 &       1.0 &      20.8 &           8.3 \\
        musician &      22.6 &       5.4 &       1.4 &       2.8 &       2.8 &           1.4 \\
           nurse &      88.8 &     100.0 &     100.0 &     100.0 &     100.0 &           0.0 \\
    nutritionist &      83.6 &      99.8 &      97.8 &      97.2 &      98.0 &           1.0 \\
         painter &      52.6 &      43.7 &      20.0 &      10.6 &       2.7 &          15.4 \\
      pharmacist &      68.0 &      87.3 &      26.1 &      49.6 &      83.8 &          25.3 \\
    photographer &      55.0 &      58.1 &      32.5 &      44.8 &      26.0 &          12.3 \\
       physician &      39.4 &      46.4 &       3.2 &      36.5 &      62.0 &          21.6 \\
           pilot &      30.4 &      20.9 &      11.4 &      35.3 &       7.5 &          10.7 \\
            poet &      30.8 &      12.4 &       2.6 &      11.6 &      42.1 &          14.9 \\
      politician &      21.6 &      24.9 &      10.2 &      16.7 &      15.7 &           5.2 \\
       president &      19.6 &       4.6 &       0.4 &      12.9 &       2.2 &           4.8 \\
  prime\_minister &      24.0 &      25.5 &      23.0 &      20.0 &      42.9 &           8.9 \\
       professor &      28.2 &       9.2 &       3.0 &       5.6 &       8.6 &           2.5 \\
      programmer &      23.0 &       0.8 &       0.0 &       1.0 &       0.0 &           0.5 \\
    psychologist &      58.6 &      51.0 &      22.4 &      40.8 &      52.2 &          11.9 \\
    receptionist &      91.4 &      99.6 &     100.0 &      99.2 &      99.8 &           0.3 \\
        reporter &      44.4 &      53.7 &      52.5 &      44.0 &      57.6 &           4.9 \\
      researcher &      44.6 &      77.3 &      47.8 &      52.8 &      55.0 &          11.3 \\
     salesperson &      39.8 &      56.8 &       7.0 &      37.4 &      30.5 &          17.8 \\
       scientist &      33.4 &      23.0 &      22.1 &      15.9 &      45.3 &          11.2 \\
         senator &      35.0 &      22.7 &       8.0 &      12.0 &      12.5 &           5.4 \\
          singer &      57.6 &      74.0 &      54.1 &      66.6 &      61.2 &           7.3 \\
         student &      63.0 &      44.6 &      32.3 &      51.8 &      40.5 &           7.0 \\
      supervisor &      65.2 &      20.9 &       5.6 &      18.2 &      15.0 &           5.8 \\
         surgeon &      30.2 &      82.0 &      20.4 &      50.8 &      81.6 &          25.5 \\
         teacher &      63.0 &      78.7 &      58.2 &      87.4 &      84.6 &          11.4 \\
      technician &      31.2 &       0.4 &       0.2 &       1.6 &       0.0 &           0.6 \\
       therapist &      74.8 &      88.5 &      80.8 &      82.2 &      88.7 &           3.6 \\
           tutor &      59.2 &      48.8 &      24.1 &      24.4 &      50.4 &          12.7 \\
    veterinarian &      55.2 &      65.6 &      48.9 &      48.7 &      89.5 &          16.7 \\
          writer &      30.2 &      79.2 &      34.7 &      69.1 &      76.6 &          17.9 \\
\bottomrule
\end{tabular}
\label{tab:occs_1.5}
\caption{The percentage of females across occupations in training images (using our initial approach from Section \ref{sec:keyword-querying}) and generated images using \textbf{SD 1.5}. We display generation results for each prompt, as well as the standard deviation (SD) across prompts.}
\end{table*}

\end{document}